\documentclass{article}

\usepackage{arxiv}

\usepackage[utf8]{inputenc} 
\usepackage[T1]{fontenc}    
\usepackage{hyperref}       
\usepackage{url}            
\usepackage{booktabs}       
\usepackage{amsfonts}       
\usepackage{nicefrac}       
\usepackage{microtype}      
\usepackage{lipsum}
\usepackage{graphicx}
\graphicspath{ {./images/} }

\usepackage{xspace}
\usepackage{multicol}
\usepackage{multirow}
\usepackage{makecell}
\usepackage{arydshln}
\usepackage{algorithm}
\usepackage{algorithmic}
\usepackage{amssymb}%
\usepackage{pifont}
\usepackage{amsmath}

\usepackage{xcolor}
\usepackage{subcaption}
\usepackage{dcolumn}
\usepackage{enumitem}
\usepackage{titlesec}

\title{Weak Supervision for Real World Graphs}

\author{
Pratheeksha Nair \\
  School of Computer Science\\
  McGill University\\
  Montreal, QC \\
  \texttt{pratheeksha.nair@mail.mcgill.ca} \\
   \And
 Reihaneh Rabbany \\
  School of Computer Science\\
  McGill University\\
  Montreal, QC \\
  \texttt{rrabba@cs.mcgill.ca} \\
}
\date{}

\newcommand{\method}{\textsc{WSNet}\xspace}
\newcommand{\htusa}{\textsc{ASW-Real}\xspace}
\newcommand{\htgen}{\textsc{ASW-Synth}\xspace}
\newcommand{\misinfo}{\textsc{Liar-WS}\xspace}
\newcommand{\cora}{\textsc{Cora-WS}\xspace}

\begin{document}
\maketitle
\begin{abstract}
Node classification in real-world graphs often suffers from label scarcity and noise, especially in high-stakes domains like human trafficking detection and misinformation monitoring. While direct supervision is limited, such graphs frequently contain weak signals—noisy or indirect cues—that can still inform learning. We propose \method, a novel weakly supervised graph contrastive learning framework that leverages these weak signals to guide robust representation learning.\method integrates graph structure, node features, and multiple noisy supervision sources through a contrastive objective tailored for weakly labeled data. Across three real-world datasets and synthetic benchmarks with controlled noise, \method consistently outperforms state-of-the-art contrastive and noisy-label learning methods by up to 15\% in F1 score. Our results highlight the effectiveness of contrastive learning under weak supervision and the promise of exploiting imperfect labels in graph-based settings.
\end{abstract}

\section{Introduction}
Despite impressive advancements in graph neural networks (GNNs)~\cite{kipf2017semisupervised,velivckovic2017graph} for learning effective node representations, their dependence on large, cleanly labeled datasets remains a critical limitation~\cite{sun2020multi,zhou2019effective}.  In many high-stakes applications, obtaining reliable labels at scale is not only challenging but often impractical. Consider organized crime detection within online networks~\cite{tnet}. Obtaining labels as to whether particular actors (e.g individuals, pieces of evidence, organizations, etc.) are linked to criminal activity involves case-building and investigation which can take months, if not years. This makes it challenging to obtain labeled datasets for solving this problem through data-driven techniques. Another example is misinformation detection. High-quality training data for this task is limited because accurately labeling misinformation requires expert human review, which is both time-consuming and subjective. Relying on inexpert human annotators to scale data collection can introduce biased labels~\cite{allen2021scaling} and using algorithms to scale key parts of misinformation detection can have important justice implications and shape fact-checking verdicts~\cite{Neumann2022justice}.

Recently, machine learning practitioners have been turning to programmatic weak supervision (PWS)~\cite{ratner2016} where large amounts of noisy, incomplete or imprecise labels are cheaply developed from various sources such as distant supervision, external knowledge bases, or heuristics/rules~\cite{wu2022learning}. These weak signals can be aggregated to approximate true labels. This approach has been shown to be particularly valuable in cases where traditional labeling is expensive or difficult to acquire; such as developing biomedical datasets~\cite{fries2022bigbio}, cyber-crime detection~\cite{tnet}, and resonant anomaly detection in high-energy physics~\cite{chen2023resonant}. 

Another promising direction is self-supervised learning (SSL), particularly graph contrastive learning (GCL)\cite{velivckovic2018deep}, which enables representation learning without labels by contrasting multiple augmented views of a graph. Building on SSL, supervised contrastive learning (SupCon)\cite{supcon} has shown that aligning representations according to class labels significantly enhances performance. Recently, this approach has been extended to graphs in the form of ClusterSCL through graph structure based cluster-aware data augmentation~\cite{clusterscl}. 

Our work introduces \method, a novel weakly supervised graph contrastive learning framework that lies at the intersection of PWS and GCL. \method is designed to effectively leverage signals from both weak labels and contrastive learning to learn high-quality node representations that align with class labels. A key challenge in weakly supervised learning is that the weak labels may distort the learned representations by incorrectly pulling apart nodes in the embedding space with the same inherent labels. To counter this, \method integrates a regularization term informed by the graph structure based on contrastive learning. By combining both the contrastive and classification objectives, \method learns representations that are better aligned with their class labels.

In summary, our main contributions are:
\setlist{nolistsep}
\begin{enumerate}[noitemsep]
    \item \textbf{Weak supervision for real-world graphs}: We demonstrate that weak supervision can efficiently generate large amounts of imperfect labels for real-world graph nodes, particularly when traditional, high-quality labels are difficult or expensive to obtain. 
    \item \textbf{Novel methodology}: We present a new method called \method, which combines weak supervision and contrastive learning for GNNs. \method addresses potential issues of weak supervision by incorporating a contrastive loss term that aligns node embeddings with class labels. Code for \method is made public on \href{https://github.com/nair-p/WSNET-Weakly_Supervised_Graph_Contrastive_Learning}{Github}.
    \item \textbf{GCL on diverse domains}: 
    We validate \method on four different real-world datasets, including organized crime detection~\cite{tnet}, academic paper classification within citation networks~\cite{cora_orig}, and misinformation detection~\cite{liar}, demonstrating its generalizability. Our results supports that \method can be effectively adapted to multiple domains, expanding its impact.
\end{enumerate}

\section{Related Works}
Here, we position weakly supervised graph contrastive learning in the context of four related fields. 

\paragraph{Programmatic Weak Supervision (PWS):}

In this paradigm, multiple weak supervision sources such as heuristics, knowledge bases, pre-trained models, etc are encoded into \textit{labeling functions} (LFs) that provide labels for a subset of the data~\cite{survey}. The LFs may be noisy, erroneous and, provide conflicting labels. To address this, label models were developed that \textit{aggregate} the noisy votes of the LFs to obtain training labels which are then used to train models for downstream tasks~\cite{ratner2016, ratner2019, varma2019b, fu2020}. 
Our work is related to PWS in that it utilizes multiple weak labels but our focus is not on label aggregation. We simply use the signals from the weak labels to improve the contrastive learning process. We use majority vote (MV), the simplest and most straightforward strategy for label aggregation, that chooses a label based on the majority consensus of all LFs. Other approaches which were designed to consider input features for a classification task either can not directly be applied to graphs, rely on pre-trained language models, or require additional inputs such as error rates of LFs or a set of labelled data~\cite{survey}. We do not compete with these approaches as our focus is not on weak label aggregation or label denoising but rather, studying its effects on GCL. 

\paragraph{Noisy Label Learning:}

Weak labels that are aggregated using majority vote are still incompletely accurate and noisy. This setting is  different from noisy label learning (NLL) where each node has only one noisy label. In these cases, most solutions are focused on denoising the labels and/or loss regularization~\cite{nrgnn, pignn}. There have also been early efforts on incorporating GCL for noise robust learning~\cite{yuan2023learning}. 
PI-GNN~\cite{pignn} is a recent work that introduces an adaptive noise estimation technique leveraging pairwise interactions between nodes for model regularization. NRGNN~\cite{nrgnn} is another recent work that utilizes edge prediction to predict links between unlabelled and labelled nodes and expands the training set with pseudo labels, making it more robust to label noise. Unlike these works, our method does not handle or `clean' label noise but rather uses weak label signals to improve the node representation learning. 

\paragraph{Graph Contrastive Learning:}
Contrastive learning focuses on pulling a node and its \textit{positive} sample closer to each other in the embedding space, while pushing it away from its \textit{negative} samples~\cite{supcon,chen2020simple}. For contrastive learning in graphs, node and graph level augmentations are often contrasted in different ways. DGI~\cite{velivckovic2018deep} contrasts graph and node embeddings within one augmented view. GraphCL~\cite{you2020graph} maximizes the agreement between two augmented views of the same graph. MVGRL~\cite{hassani2020contrastive} augments the graph using node diffusion whereas GRACE~\cite{zhu2020deep} augments graph views using edge removal and feature masking. CSGCL~\cite{csgcl} uses graph augmentations based on community strength and structure while GCA~\cite{zhu2021graph} uses adaptive augmentations based on topological and semantic graph properties. 
BGRL~\cite{thakoor2021large} predicts alternative augmentations for the nodes and alleviates the need for negative contrast pairs. GMI~\cite{peng2020graph} formally generalizes mutual information for the graph domain. SUGRL~\cite{Mo_AAAI_2022} complements structural and neighborhood information to enlarge intra-class variation without any graph augmentations. Similarly, iGCL~\cite{li2023iGCL} introduces an invariant-discriminant loss that is free from augmentations and negative samples. SelfGNN~\cite{kefato2021selfsupervised} proposes a GCL approach that uses feature augmentations over topological augmentations and does away with negative sampling. 
Our contrastive learning method differs from these approaches in one major way in that it leverages signals from weak labels. 

\paragraph{Supervised Graph Contrastive Learning:}
SupCon~\cite{supcon} is a supervised contrastive learning method for ImageNet classification and was adapted to graphs in ClusterSCL~\cite{clusterscl}. It uses class label information in the contrastive loss to learn efficient embeddings. To negate the impacts of SupCon induced by the intra-class variances and the inter-class similarities they combine it with node clustering and cluster-aware data augmentation. 
JGCL~\cite{jgcl} further incorporates both supervised and self-supervised data augmentation and propose a joint contrastive loss. None of these are directly suitable for multiple weak label learning.
~\cite{zheng2021weakly} proposed a weakly supervised contrastive learning framework for image classification and used node similarity to obtain weak labels. These weak labels are dependent on the augmented views of their graph and are inherently different from the weak/noisy class labels in our setting.
However, we do compare \method with the graph-adapted version of SupCon~\cite{clusterscl}.
Recently, ~\cite{cui2023rethinking} discussed theoretically that noisy labels do not help select ``clean labels'' for training in contrastive learning. Our work is different as firstly, we do not focus on clean label selection and secondly, our work relies on signals from \textit{multiple} weak labels. Thirdly, our method also draws on valuable information from the graph structure which we show helps learn robust node embeddings.

\section{Preliminaries}

\paragraph{\textbf{Graph notation}:}
We represent a graph with $N$ nodes by $G = (A, X)$, where $X \in \mathcal{R}^{N\times d}$ is the $d$-dimensional node feature matrix and $A \in \{0, 1\}^{N\times N}$ is the adjacency matrix. 
Each node $i$ also has an unobserved true label $y_i \in \{0, 1, ..., C-1\}$ and $Y = [y_1,...y_N]$ represents the ground truth labels

\paragraph{\textbf{Weak labels}:}
Each node $i$ has $m$ weak labels $\Lambda_i \in \{-1,0\dots C-1\}^m$. These weak labels are obtained from $m$ different sources called \textit{labeling functions} (LF), represented as $\lambda_j:\{1, \dots N\}\rightarrow \{-1,0, \dots C-1\}$ and $j\in \{1,..m\}$, that map an input node to a weak label or -1 (abstain). An LF $\lambda_j$ applied on node $i$ produces a weak label $\Lambda_{ij} \in \{-1, 0, \dots, C-1\}$.
Thus, $\Lambda_i = [\Lambda_{i1}...\Lambda_{im}] = [\lambda_1(i) \dots \lambda_m(i)]$ and $\Lambda \in \{-1, 0, \dots C-1\}^{N \times m}$ gives the weak label matrix for all $N$ nodes in the graph. 
Given $\Lambda_{i}$, an aggregated label $\tilde{y_i}$ is the most frequently appearing label in $\Lambda_i$ (i.e, majority vote).
$\tilde{Y} = [\tilde{y_1},...,\tilde{y_N}]$ represents the aggregated labels for all $N$ nodes in the graph.

\paragraph{\textbf{Problem statement}:} Given $G=(A,X)$ and $\Lambda$, the goal is to learn a graph neural network classifier $F: G \rightarrow Y$.

\section{Proposed Method: \method}
\begin{figure*}[t]
    \centering\vspace{-10pt}
    \includegraphics[width=0.7\textwidth]{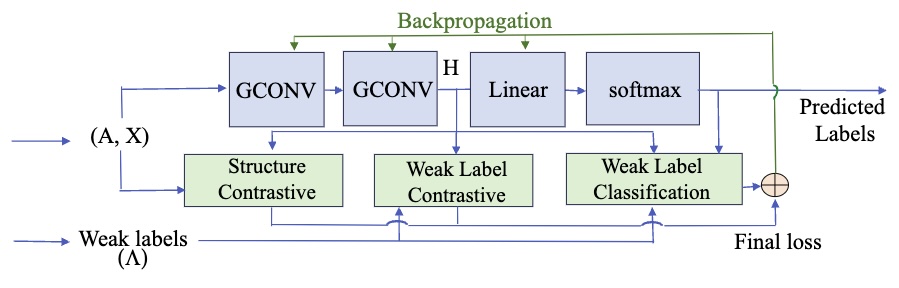}
    \caption{The graph $(A, X)$ and weak labels ($\Lambda$) are input to the \method pipeline which then produces label predictions.}
    \label{fig:method}
\end{figure*}

We introduce \method, shown in Figure \ref{fig:method}, a GNN consisting of two graph convolutional layers that learn hidden node representations $H$ which are then mapped to the output dimension space by a fully connected linear layer followed by softmax to obtain a probability distribution over all the class labels.
Drawing on ideas from information theory, \method combines contrastive learning with weak supervision by jointly optimising a three part loss function consisting of a  weak label classification component, weak label contrastive component and structure-based contrastive component.

\begin{figure}
    \centering
    \includegraphics[width=.8\linewidth]{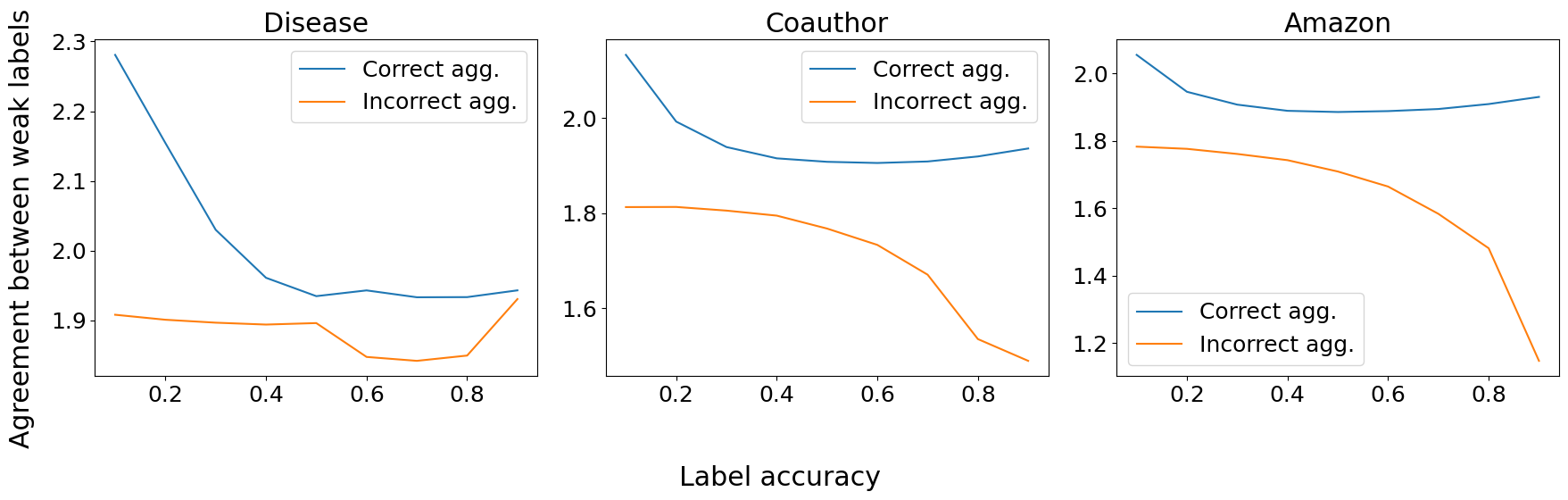}
    \caption{Nodes whose majority vote aggregated label corresponds to the ground truth (`Correct agg.') have a higher weak label agreement than `Incorrect agg.'.}
    \label{fig:lf_entropy}
\end{figure}

\paragraph{Weak Label Classification Component:} This term calculates the cross-entropy loss between predicted labels from \method ($\hat{Y}$) and majority-vote aggregated labels ($\tilde{Y}$), weighted to account for label noise. Node weights are determined by two factors: 1) the agreement among weak labels (measured by entropy) and 2) the node's representativeness in the embedding space.

Empirically (see Figure \ref{fig:lf_entropy}), nodes with low weak label entropy (higher agreement) tend to have more reliable aggregated labels, corroborating findings in prior work~\cite{tnet}. Such nodes are assigned higher weights. Conversely, nodes closer to cluster centroids (obtained using K-Means clustering with $k=C$) in embedding space, calculated via normalized cosine similarity, are considered more representative and weighted lower if their labels are uncertain. Node weights ($\rho_i$) are computed using Equation \ref{eq:rho_eqn}. 
\begin{gather}\vspace{-10pt} \label{eq:rho_eqn}
\rho_i = |Q_i| * \frac{(h_i \cdot h_{Q_i})}{\sum_{j=1}^{N} (h_j \cdot h_{Q_j})} * entropy(\Lambda_i) \vspace{-10pt} 
\end{gather}
where $h_i$ is the hidden representation of node $i$ and $h_{Q_i}$ is the centroid representation of cluster $Q_i$ that $i$ belongs to. `.' indicates cosine similarity.

The weighted weak label cross-entropy loss is then given by Equation \ref{eq:weak_label_ce_loss}. 
\begin{gather} \vspace{-10pt} \label{eq:weak_label_ce_loss} 
L_{WLCE} = \frac{1}{N} \sum_{i=1}^N \text{CrossEntropy}(\hat{Y}_i, \tilde{Y}_i) * \rho_i \vspace{-10pt} 
\end{gather}

\begin{figure}
    \centering\vspace{-10pt}
    \includegraphics[width=0.9\linewidth]{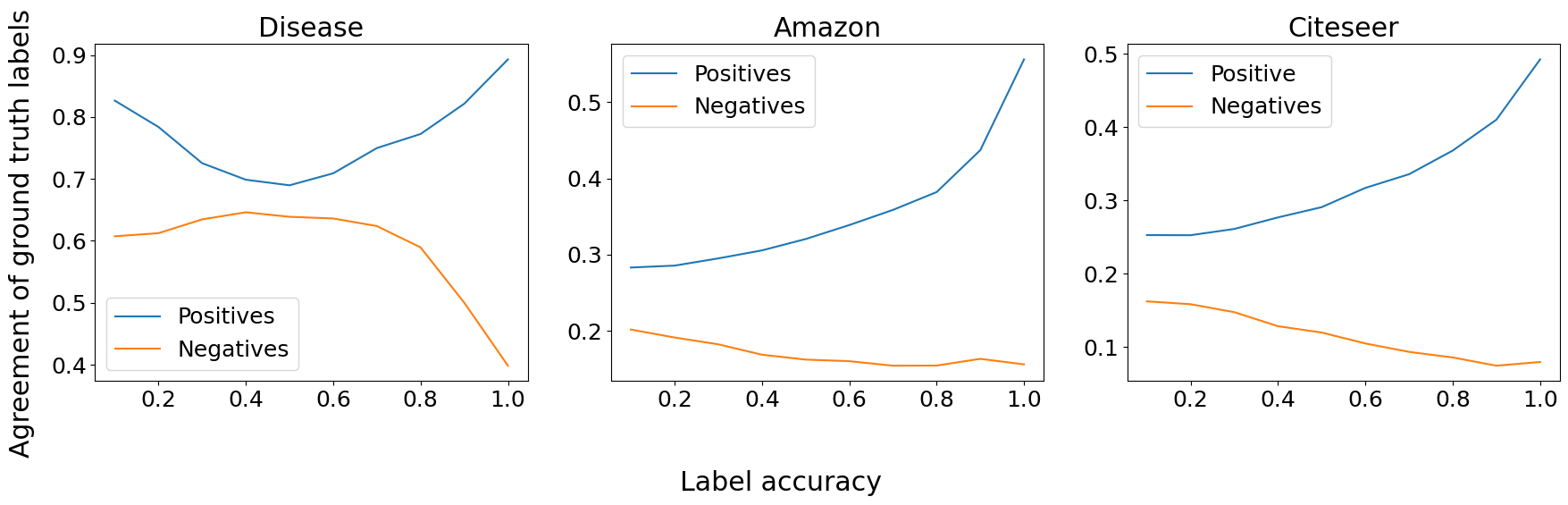}
    \caption{\small Node pairs with highly similar weak label distributions (chosen as positives) belong to the same class more often than pairs with low weak label similarity (negatives).}
    \label{fig:wl_cl}
\end{figure}

\paragraph{Weak Label Contrastive Component:}\vspace{-10pt} This component uses the InfoNCE contrastive learning loss~\cite{infonce} to align nodes with similar weak label distributions while separating those with dissimilar distributions. Nodes with similar weak label distributions—indicating identical LF voting patterns — are likely to belong to the same class, regardless of label accuracy. We show that the top 50 node pairs with the highest pairwise cosine similarity of weak labels on average belong to the same class (see Figure \ref{fig:wl_cl}). Therefore, their embeddings should be closer in the representation space.

To implement this, for each node, its positive pair is sampled based on the cosine similarity of weak label distributions. Negative pairs are created by randomly selecting nodes that are not connected to the given node. 
The contrastive loss is computed using InfoNCE loss given by Equation \ref{eq:weak_label_con_loss}. 
\begin{gather} \label{eq:weak_label_con_loss} 
L_{WLCon} = -\log{\frac{\exp{(h_i.h_i^+)}/tau}{\sum\limits_{j=1}^r\exp{(h_i.S^-_j)}/tau + \exp{(h_i.h_i^+)}/tau}}\vspace{-10pt}
\end{gather}
Here $h_i^+$ is the representation of the positive pair for node $i$ and $S_j^-$ is the set of $r$ negative samples chosen for node $i$. `.' indicates dot product. $tau$ is the temperature parameter that scales the logits before applying softmax creating more or less uniformly distributed feature spaces. 

\paragraph{Structure-based Contrastive Component:}
\method also incorporates a structure-aware contrastive learning component that maximizes mutual information between a node’s embedding and its graph community, leveraging both local and intermediate graph structures. Communities are densely connected node groups that are loosely connected to others, making them ideal for capturing mesoscale patterns. By pooling over communities instead of the entire graph, the method generates more granular global representations, improving embeddings for tasks like node classification, particularly in graphs with strong community structures, as validated by our experiments.

The structure-based contrastive loss, $L_{SCon}$, given by Equation \ref{eq:structure_con_loss}, optimizes node embeddings by contrasting them with both real and corrupted graph counterparts. This loss also acts as a regularizer, mitigating the impact of label noise. Here, $h_i$ represents the embedding of node $i$, $B_i$ is the community to which node 
$i$ belongs, $h_{B_i}$ is the mean-pooled embedding of all nodes in $B_i$, and $\tilde{h}_j$ is the corrupted embedding of node $j$, obtained by shuffling the node features.

\begin{gather} \label{eq:structure_con_loss}
\mathcal{L}_{SCon} = -\mathbb{E}_G \left[ \sum_{i=1}^{N} \log \sigma\left(\mathbf{h}_i^\top \mathbf{h}_{B_i}\right) + \sum_{j=1}^{N} \log \left(1 - \sigma\left(\tilde{\mathbf{h}}_j^\top \mathbf{h}_{B_j}\right)\right) \right]
\end{gather}

The final loss function that \method optimizes is a linear combination of the three loss components and is given by Equation \ref{eq:full_loss}.
\begin{gather} \label{eq:full_loss}
    L = L_{SCon} + L_{WLCE} + L_{WLCon}
\end{gather}

\section{Experiments}
We evaluated \method on real-world hard-to-label graphs and synthetic benchmark datasets, comparing its performance against a diverse range of baseline methods. On synthetic datasets, we demonstrated \method's robustness and consistent superiority across varying levels of label noise. For all experiments, we used five independent 80-10-10 train-validation-test splits and reported the average and standard deviation of weighted F1 scores on the test set. During training, all methods had access only to weak labels from the train split. The experiments were conducted on a local MacBook M2 with 8 CPU cores and a remote 1-core GPU.

\subsection{Baselines}
We compared \method against baselines from four categories:
\begin{itemize}
    \item \textit{Self-Supervised Learning (SSL/GCL}): Methods leveraging augmented views, node neighborhoods, and negative sampling.
    \item \textit{Noisy Label Learning (NLL)}: NRGNN\cite{nrgnn} and PI-GNN\cite{pignn}, designed for learning with noisy labels.
    \item \textit{Programmatic Weak Supervision (PWS)}: Majority Vote, Snorkel\cite{ratner2019}, and Hyper Label Model\cite{wu2022learning}.
    \item \textit{Supervised Graph Contrastive Learning (SupGCL)}: SupCon\cite{supcon} and ClusterSCL\cite{clusterscl}, focused on cluster-aware supervised learning.
\end{itemize}

We used the official implementations and default hyperparameters for all baselines. \method was trained for 200 epochs, with hyperparameters $r$ and $tau$ fine-tuned on a validation set. For GCL baselines, node embeddings were used to train a downstream logistic regression classifier using majority vote-aggregated labels. For PWS baselines, aggregated weak labels were used to train the classifier. 

\subsection{Real-world use cases}

\subsubsection*{Misinformation detection in news statements}
With advancements in generative AI, misinformation has become a growing societal concern~\cite{chen2023combating}. While Large Language Models (LLMs) are effective in detecting misinformation~\cite{chen2023combating,pelrine2023towards}, they suffer from hallucination and outdated knowledge. To address this, Retrieval-Augmented Generation (RAG) integrates LLMs with external data sources, enabling verification through queries grounded in recent information.
We developed \misinfo, a graph-based resource built on the LIAR dataset~\cite{liar}, which contains 12.8k annotated news statements with true/false labels from PolitiFact.com.
Using ChatGPT-3.5-turbo API, we generated 10 paraphrases for each statement to simulate variations in reporting. Original statements and their paraphrases are nodes that are connected via edges. Text embeddings were computed as node features using Cohere LLM for semantic representation. Weak labels were derived from misinformation predictions by GPT-4-0125 and Claude 3-Haiku, using RAG with Cohere Search, DuckDuckGo, and a no-search baseline, yielding six weak label sources.
The LLM-driven paraphrasing introduces semantic variability; the RAG predictions and multiple weak labeling sources produces noisy and inconsistent labels combined with ambiguity in defining misinformation makes \misinfo a valuable dataset for future research in this domain.

\subsubsection*{Detecting organized activities from online escort advertisements}

Human trafficking (HT) for sexual exploitation remains a widespread issue globally. Traffickers often advertise victims’ services on adult service websites (ASWs), which host a mix of ads, including independent escorts, agency-run posts, and spam. Detecting suspicious activity in these ads is challenging due to their sensitive, complex, and hard-to-label nature.

The problem was framed as a weakly supervised graph learning task~\cite{tnet}, focusing on a 3-class classification: human trafficking, spam, and independent escorts. The graph connects groups of escort ads through shared contact information, such as phone numbers or email addresses. Node features include cluster-level statistics, such as cluster size, counts of phone numbers and URLs. Domain experts defined 12 LFs, including the presence of suspicious keywords and indicators of multiple individuals advertised in an ad.
Two datasets -- 1) \htusa: A real-world dataset of escort ads and 2)\htgen: A synthetic dataset of realistic escort ads, facilitate studying HT detection, offering real-world and controlled data environments for testing graph-based learning approaches.

\subsubsection*{Research paper topic classification}
We introduce \cora, a weakly supervised extension of the benchmark Cora dataset, designed for classifying academic papers into research topics within the field of Machine Learning in Computer Science~\cite{cora_orig}. The dataset represents research papers as nodes, connected by edges if they cite each other. Each node is characterized by TF-IDF vectorized representations of its abstract, providing rich textual features for learning tasks.

To generate weak labels, we manually curated a list of relevant keywords for each class and defined two rule-based LFs that strictly match these keywords in the 1) abstract and 2) title of the paper. For instance, the class "Reinforcement Learning" includes keywords such as reward, action, state, and environment. If a paper’s abstract or title contained these keywords, the corresponding LF assigns it to this class. 
Irrelevant context, missing keywords and limited semantic understanding which overlooks nuanced relationships may result in incorrect labels. Additionally, imbalanced label distributions and the reliance on manually curated domain-specific keywords further complicate learning, making \cora a valuable benchmark for testing weak supervision robustness.


More dataset statistics are provided in Table \ref{tab:summary_real_world}. Through these datasets, our study demonstrates \method's adaptability across diverse real-world scenarios

\begin{table}[]
\centering
\resizebox{0.9\linewidth}{!}{
\begin{tabular}{@{}lllrrrcccc@{}}
\toprule
Type                                                                             & Dataset   & $N$   & Deg. & |Comms.| & $C$ & Mean Acc.     & Max Acc.              & Cov.                  & $m$                 \\ \midrule
\multirow{4}{*}{Real-world}                                                      & \misinfo  & 6468  & 1.8   & 588      &  2   & 79.9    &  87.0             & 98.8                  & 6                   \\
 & \htusa    & 438   & 398.0  & 13            & 3   & 11.0 &      13.7              & 16.4                  & 12                  \\
 & \htgen    & 7216  & 70.2   & 1571            & 3   & 10.1 & 29.9                   & 8.2                  & 12                  \\
 & \cora     & 4607  & 5.2    & 1046          & 7   & 11.1  & 14.6                 & 19.9                  & 14                  \\ \midrule
\multirow{5}{*}{Benchmark} & Citeseer  & 4230  & 2.52   & 537          & 6   & \multirow{5}{*}{10-100} & \multirow{5}{*}{10-100} &\multirow{5}{*}{70.0} & \multirow{5}{*}{10} \\
& Disease    & 1044 & 1.99   & 36           & 2   &    &                    &                       &                    \\
 & Coauthor     & 34493   & 14.38   & 22         & 5   &    &                    &                       &                     \\
& Ogbn-arxiv & 169343   & 13.7   & 142         & 40   &    &                    &                       &                     \\ 
& Amazon & 13752 & 35.8 & 332  & 10 & & & &\\\bottomrule
\end{tabular}}
\caption{Dataset Statistics. The mean and maximum accuracy of the weak labels, mean coverage (Cov.) and number of weak labels ($m$) in the real-world graphs are reported. The benchmark datasets have synthetic weak labels generated according to varying accuracies ranging from 10\% to 100\%. $N$ is the number of nodes in the graph, Deg. represents the average degree of the nodes in the graph. |Comms.| is the number of detected communities and $C$ is the number of classes.}
\label{tab:summary_real_world}
\end{table}

\subsection{Synthetic Experiments} 

The synthetic experiments simulate our method's performance for varying noise levels. For each dataset, we created $m$ weak labels by randomly flipping the ground-truth label according to a noise ratio ($1 - p_a$). We varied the label accuracy $p_a$ from 0.1 \dots 1 (i.e, noise ratio 0.9 \dots 0) keeping coverage fixed at 70\% (i.e, 30\% of the data samples have weak label -1 or abstain). 

We ran our experiments on five benchmark node classification datasets of varying sizes (see Table \ref{tab:summary_real_world}).
Citeseer\cite{citeseer} is a citation network with nodes as research papers, edges as citations, and TF-IDF node attributes, used to classify papers into six topics. Disease\cite{disease} is another citation network for biomedical papers, with classification based on research topics. ogbn-arxiv~\cite{mag} is another large citation network of CS arXiv papers (nodes) to be classified into one of 40 subject areas.
Amazon Product\cite{amazon} is a co-purchase graph for predicting product categories or recommendations using review embeddings or metadata. Coauthor Physics\cite{coauthor} is a co-authorship network, where author embeddings classify researchers into physics-related fields.

\section{Results}

\begin{table*}[]
\centering
\resizebox{0.9\linewidth}{!}{
\begin{tabular}{@{}lccccr@{}}
\toprule
     & \htgen         & \cora           & \htusa          & \misinfo               & Method          \\ \midrule
DGI~\cite{velivckovic2018deep}        & 0.20 $\pm$ 0.07          & 0.23 $\pm$ 0.01          & 0.36 $\pm$ 0.10          & 0.83 $\pm$ 0.01          & \multirow{6}{*}{SSL}    \\
GRACE~\cite{zhu2020deep}      & 0.28 $\pm$ 0.09          & 0.32 $\pm$ 0.03        & 0.52 $\pm$ 0.06          & 0.84 $\pm$ 0.01          &                         \\
MVGRL~\cite{hassani2020contrastive}      & 0.52 $\pm$ 0.01         & 0.22 $\pm$ 0.09          & 0.53 $\pm$ 0.08          & 0.84 $\pm$ 0.01          &                         \\
GCA~\cite{zhu2021graph}        & 0.19 $\pm$ 0.06          & 0.24 $\pm$ 0.09          & \textcolor{blue}{0.56 $\pm$ 0.10 }         & 0.82 $\pm$ 0.02          &                         \\
BGRL~\cite{thakoor2021large}       & 0.28 $\pm$ 0.01          & 0.31 $\pm$ 0.03          & 0.53 $\pm$ 0.08          & 0.81 $\pm$ 0.01          &                         \\
SUGRL~\cite{Mo_AAAI_2022}      & 0.35 $\pm$ 0.02          & 0.32 $\pm$ 0.07          & 0.42 $\pm$ 0.08          & 0.84 $\pm$ 0.01          &                         \\  \cdashline{1-6}
Random & 0.35 $\pm$ 0.00 & 0.15 $\pm$ 0.002 & 0.40 $\pm$ 0.01 & 0.57 $\pm$ 0.003 & \\
Majority Vote & \textcolor{blue}{0.58 $\pm$ 0.001} & 0.30 $\pm$ 0.001 & 0.13 $\pm$ 0.01 & 0.81 $\pm$ 0.001 & \multirow{2}{*}{PWS}\\
Snorkel\cite{ratner2019} & \textcolor{blue}{0.58 $\pm$ 0.00} & 0.22 $\pm$ 0.003 & 0.10 $\pm$ 0.09 & \textcolor{blue}{0.86 $\pm$ 0.09} & \\
Hyper Label Model\cite{wu2022learning} & 0.35 $\pm$ 0.00 & \textcolor{blue}{0.33 $\pm$ 0.00} & 0.23 $\pm$ 0.09 & 0.83 $\pm$ 0.00 & \\ \cdashline{1-6}
NRGNN~\cite{nrgnn}      & 0.55 $\pm$ 0.15          & 0.10 $\pm$ 0.01          & 0.32 $\pm$ 0.13          & 0.77 $\pm$ 0.04          & \multirow{2}{*}{NLL}    \\
PI-GNN~\cite{pignn}     & 0.52 $\pm$ 0.01          & 0.32 $\pm$ 0.06          & 0.42 $\pm$ 0.08          & 0.77 $\pm$ 0.01          &                         \\ \cdashline{1-6} 
SupCon~\cite{supcon}     & 0.52 $\pm$ 0.03          & 0.18 $\pm$ 0.02          & 0.42 $\pm$ 0.06          & 0.77 $\pm$ 0.02          & \multirow{2}{*}{SupGCL} \\
ClusterSCL~\cite{clusterscl} & 0.22 $\pm$ 0.08          & 0.13 $\pm$ 0.06          & 0.28 $\pm$ 0.07          & 0.84 $\pm$ 0.01         &                         \\ \cdashline{1-6} 
Ours       & \textbf{0.78 $\pm$ 0.08} & \textbf{0.40 $\pm$ 0.05} & \textbf{0.72 $\pm$ 0.03} & \textbf{0.88 $\pm$ 0.01} & \method  \\ 
\bottomrule 
\end{tabular}}
\caption{Performance of \method on real-world datasets. The test weighted F1 classification score averaged across 5-folds. shows that \method outperforms all baselines. Second best results are in blue.} \label{tab:results_real_world}
\end{table*}

\subsection{Real-world graphs}
The weighted F1 classification scores on the real-world graph datasets are presented in Table \ref{tab:results_real_world}, demonstrating that \method consistently outperforms all baselines across all four datasets. Notably, on the \htgen and \htusa datasets, \method achieves competitive performance, scoring 78\% and 74\%, respectively, which is either on par with or closely approaches the domain-specific state-of-the-art results reported in \cite{tnet}. This highlights \method's ability to generalize well, even in challenging, domain-specific scenarios.

For the \cora dataset, where no previous benchmarks exist specifically for weak supervision, \method establishes a strong baseline, showcasing its capability to perform robustly in previously unexplored weakly supervised settings.  On the \misinfo dataset, \method achieves a misinformation detection performance of 88\%, surpassing the highest weak label accuracy (87\% in Table \ref{tab:summary_real_world}). Overall, \method's classification performance is far better than the accuracy of the weak labels alone. 
This demonstrates \method’s effectiveness in integrating diverse, noisy weak labels to produce a more accurate and unified output. 

\begin{figure}
    \centering
\includegraphics[width=\linewidth]{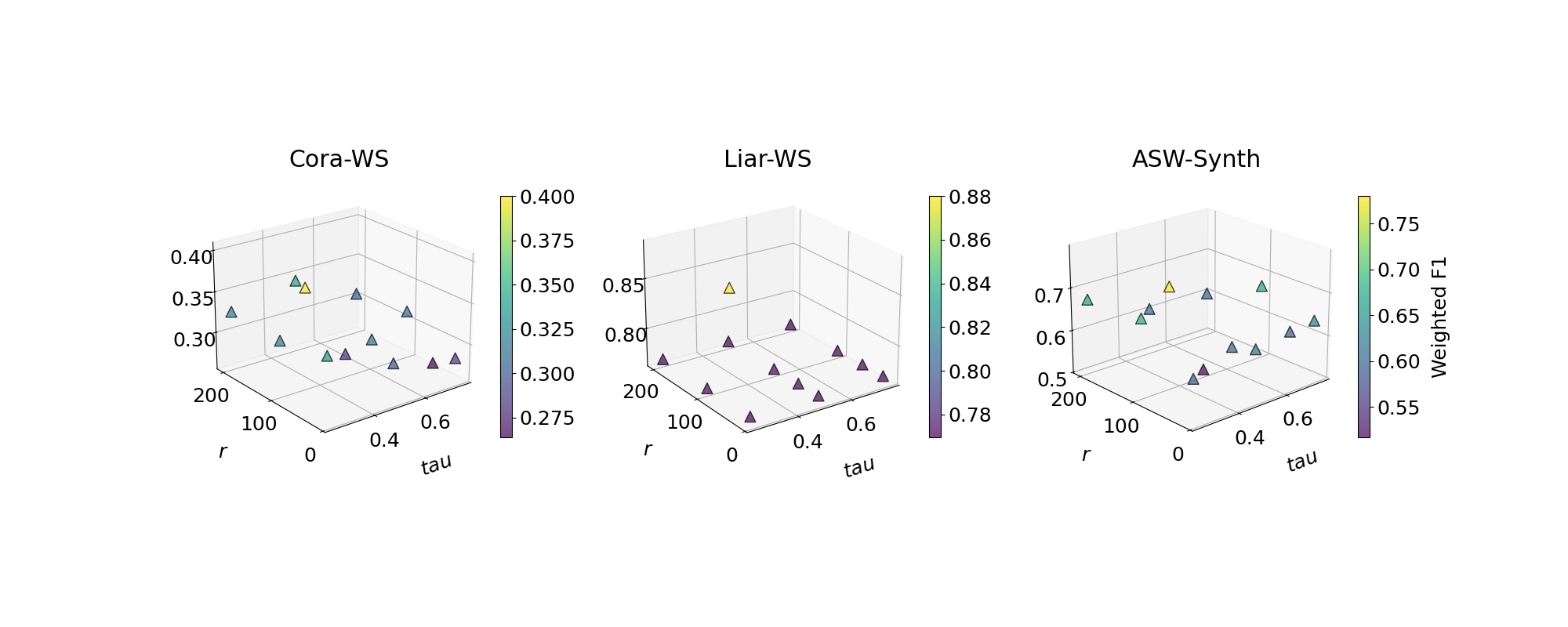}
    \caption{We fine-tuned $tau$ and $r$ on a validation set to choose the optimal hyperparameter values for each dataset.}
    \label{fig:hparam_results}
\end{figure}

To optimize performance, hyperparameters $tau$ and $r$ were fine-tuned on a held-out validation set. The results of this hyperparameter optimization are visualized in Figure \ref{fig:hparam_results}, showing that careful tuning significantly enhances performance across datasets. We see that $tau$ values closer to $0.5$ and moderate
$r$ close to $50$ typically yield the best results, balancing contrastive loss effectiveness and model stability.

\subsection{Synthetic Experiments}

\begin{figure}
    \centering
    \includegraphics[width=\linewidth]{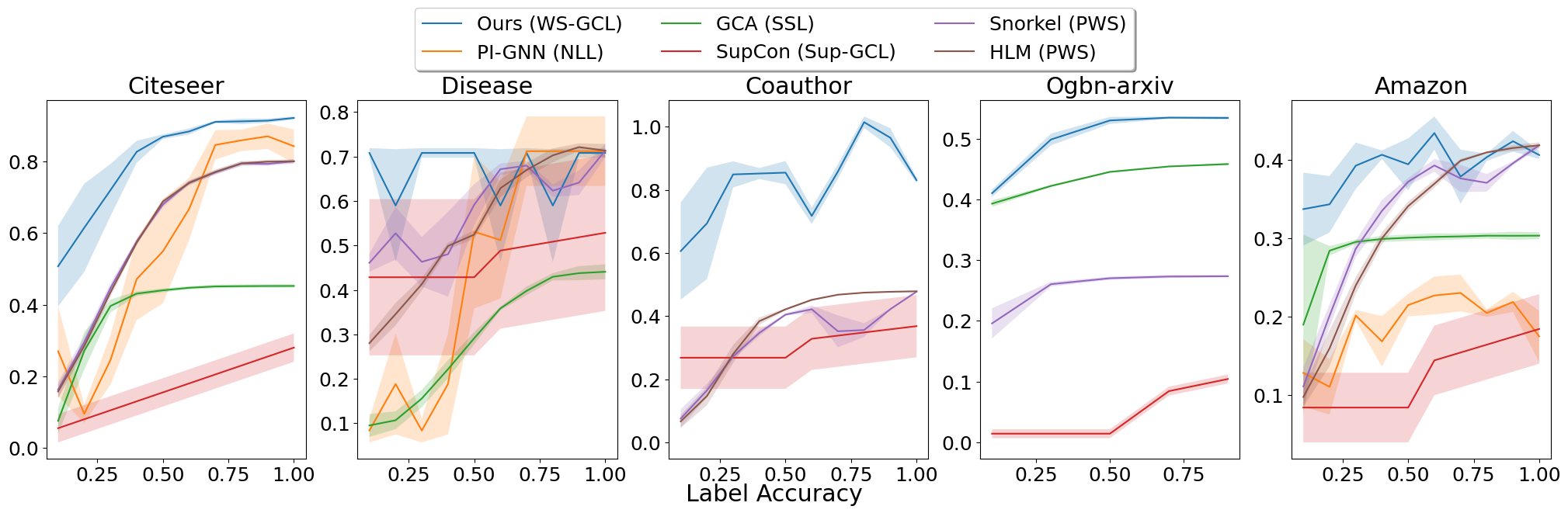}
    \caption{Performance of \method on benchmark graph datasets. The y-axis is the weighted F1 score and the x-axis is the label accuracy. \method has a clear advantage compared to baselines particularly when label accuracy is low. SupCon and PI-GNN resulted in OOM for Coauthor and HLM and PI-GNN ran OOM for ogbn-ArXiv.}
    \label{fig:synth_exp_results}
\end{figure}

Figure \ref{fig:synth_exp_results} demonstrates that \method consistently outperforms all baselines across synthetic datasets, particularly under low label accuracy conditions, showcasing its robustness in handling noisy labels. While PWS baselines perform well with high label accuracy, their performance deteriorates under noise, highlighting \method's advantage in leveraging graph structure for denoising. On Coauthor, NLL and SupGCL resulted in OOM and on Ogbn-arxiv, NLL and HLM resulted in out-of-memory issues. \method is effective in integrating multiple weak label signals with structural information, unlike NLL and SupGCL methods that focus solely on label or structural noise. This advantage highlights WS-GCL's ability to denoise and leverage imperfect supervision, offering a significant improvement over baselines in scenarios with high label noise and complex graph structures.
These results validate \method’s superior performance, scalability, and adaptability, making it highly effective for weakly supervised learning in graph tasks.

\subsection{Ablation Study}
The ablation study in Table \ref{tab:ablation_results} emphasizes the importance of each component in \method's loss function. Removing the weak label cross-entropy loss causes the most significant performance drop, highlighting the value of even noisy supervision in guiding the learning process. While the weak label contrastive loss has a smaller individual impact, it improves performance by aligning representations based on weak label agreements. The structure-based contrastive loss further enhances embeddings by capturing local and global graph structures. Together, these components integrate complementary objectives, resulting in robust, generalizable node embeddings across diverse graph datasets.

\begin{table*}[]
\centering
\resizebox{!}{0.1\linewidth}{
\begin{tabular}{@{}lccccccr@{}}

\toprule
                       & -$L_{WLCon}$ & -$L_{WLCE}$ & -$L_{SCon}$ & +$L_{WLCon}$ & +$L_{WLCE}$ & +$L_{SCon}$ & \method \\  \midrule
\misinfo                & 0.77         & 0.54        & 0.74        & 0.34                 & 0.70                  & 0.50                  & \textbf{0.88}    \\
\cora                & 0.35         & 0.15        & 0.28        & 0.10                 & 0.25                  & 0.12                  & \textbf{0.40}    \\
Citeseer ($p_c = 0.4$) & 0.71         & 0.34        & 0.62        & 0.17                 & 0.60                  & 0.20                  & \textbf{0.83}   \\
Citeseer ($p_c = 0.6$) & 0.81         & 0.52        & 0.60        & 0.15                 & 0.45                  & 0.20                  & \textbf{0.88}    \\
Disease ($p_c = 0.4$)  & 0.58         & 0.49        & 0.46        & 0.35                 & 0.51                  & 0.37                  & \textbf{0.71}    \\
Disease ($p_c = 0.6$)  & 0.70         & 0.58        & 0.68        & 0.56                 & 0.50                  & 0.40                  & \textbf{0.72}    \\ \bottomrule
\end{tabular}}
\caption{Ablation studies showing that each component of \method contributes to its performance. - indicates that loss component was removed and + indicates only that loss component was included in the final loss function.}
\label{tab:ablation_results}
\end{table*}

\section{Conclusions, Limitations and Future Work}
We introduced a graph contrastive learning method for weakly labeling complex real-world graphs and introduced two new real-world graph datasets. \method was rigorously evaluated on four real-world datasets and five benchmark graph datasets and it consistently outperformed all baselines. 
We show that combining weak label and 
structure-based contrastive learning  enhances \method's ability to learn node representation in the presence of weak labels.

We acknowledge some of our method's assumptions which warrant further investigation and analysis. For example, the impact of our methodology on data where community structure is less relevant, such as irregular biological graphs is to be investigated. The alignment between underlying class structure and weak label informed node embedding clusters can also be improved by techniques such as semi-supervised or constrained clustering.   

Nevertheless, by seamlessly integrating weak supervision with graph contrastive learning, \method bridges the gap between two distinct machine learning paradigms and is a strong proof of concept. Such a strategy has broader implications, potentially inspiring future research in graph ML for tackling noisy and complex real-world datasets. By demonstrating that weak labels can be effectively incorporated into contrastive learning frameworks, \method opens new pathways for GNN research, encouraging the development of models capable of handling diverse and imperfectly labeled data.

\section{Ethical Implications}

There are many potential societal consequences of our work, some of which have been highlighted in the paper. It must be ensured that this work is applied in the most ethical and responsible manner to any and all real-world data/applications. For example, the labeling functions defined need to inclusive, cross-verified by multiple experts and checked for encoded biases, especially if text or image based. They need to be designed in an inclusive manner by bringing into the loop, those that are most affected by it. 
For detecting organized crime activities, using labeling functions based on keywords or text embeddings can bring bias into the weak labels. Although these labeling functions are created with the help of domain experts, it is important to try and disallow the human biases from creeping into them. Our work is preliminary and studies the impact of weak labels on learned node representations and future works focused on weak label aggregation also need to be checked for bias towards particular labeling functions. 

\bibliographystyle{unsrt}  
\bibliography{references}  

\end{document}